\documentclass[runningheads]{llncs}
\usepackage{graphicx}
\usepackage{amsmath,bm}
\usepackage{threeparttable}
\usepackage{multirow}
\usepackage{adjustbox}
\usepackage{array}
\usepackage{algorithm} 
\usepackage{algpseudocode}
\usepackage{subfigure}
\usepackage{booktabs}
\usepackage{multicol}
\usepackage{amssymb}
\usepackage{dsfont}
\usepackage{algorithmicx}
\usepackage{capt-of}
\begin{document}
\title{Blending Pruning Criteria for Convolutional Neural Networks}

\author{Wei He$^{1*}$ , Zhongzhan Huang$^{2}$\thanks{Equal contribution.}, Mingfu Liang$^{3}$, Senwei Liang$^{4}$ and Haizhao Yang$^{4}$% <-this % stops a space
%\thanks{Equal contribution.}% <-this % stops a space
}

% \author{Wei He\inst{1} \and
% Zhongzhan Huang^*\inst{2} \and
% Mingfu Liang\inst{3}\and  Senwei Liang\inst{4} \and \\Haizhao Yang\inst{4}}
%

\authorrunning{Wei He et al.}
\institute{$^{1}$ Nanyang Technological University
$^{2}$ Tsinghua University \\
$^{3}$ Northwestern University $^{4}$ Purdue University\\
\email{yang1863@purdue.edu}}

\maketitle              % typeset the header of the contribution
\begin{abstract}
The advancement of convolutional neural networks~(CNNs) on various vision applications has attracted lots of attention. Yet the majority of CNNs are unable to satisfy the strict requirement for real-world deployment. To overcome this, the recent popular network pruning is an effective method to reduce the redundancy of the models. However, the ranking of filters according to their ``importance'' on different pruning criteria may be inconsistent. One filter could be important according to a certain criterion, while it is unnecessary according to another one, which indicates that each criterion is only a partial view of the comprehensive ``importance". From this motivation, we propose a novel framework to integrate the existing filter pruning criteria by exploring the criteria diversity. The proposed framework contains two stages: Criteria Clustering and Filters Importance Calibration. First, we condense the pruning criteria via layerwise clustering based on the rank of ``importance" score. Second, within each cluster, we propose a calibration factor to adjust their significance for each selected blending candidates and search for the optimal blending criterion via Evolutionary Algorithm. Quantitative results on the CIFAR-100 and ImageNet benchmarks show that our framework outperforms the state-of-the-art baselines, regrading to the compact model performance after pruning.

\end{abstract}
\section{Introduction}
Deep convolutional neural networks~(CNNs) have been the prevailing methods in computer vision and brought remarkable improvement to various tasks~\cite{he2016deep,schroff2015facenet,he2017mask,huang2019dianet,liang2020instance,huang2020efficient,liang2021drop}. However, as the CNNs are normally over-parameterized and cumbersome, it is challenging for the model deployment on devices with limited resources, and the acceleration during the inference stage becomes necessary. Network pruning, one of the critical directions in network compression, aims at eliminating the unimportant parameters or operations without compromising the model performance. In this area, filter pruning methods~\cite{li2016pruning,he2017channel,He_2021_WACV} are more practical to deploy and easily to be implemented. Generally, the workflow of filter pruning can be divided into three steps:
(1)~\textit{Normal Training}: train the original network on a specific dataset from scratch.
(2)~\textit{Pruning}: prune the insignificant network components such as neurons, filters, based on a well-handcrafted criterion, where the score magnitude under the criterion reflects their ``importance''.
(3)~\textit{Finetuning}: recover the performance loss caused by the removal of the components to a certain extent.
Among all steps, an effective pruning criterion plays an essential role in one filter pruning algorithm.

\begin{figure*}[tbp]
% \begin{center}
\centering

\includegraphics[width=0.8\textwidth]{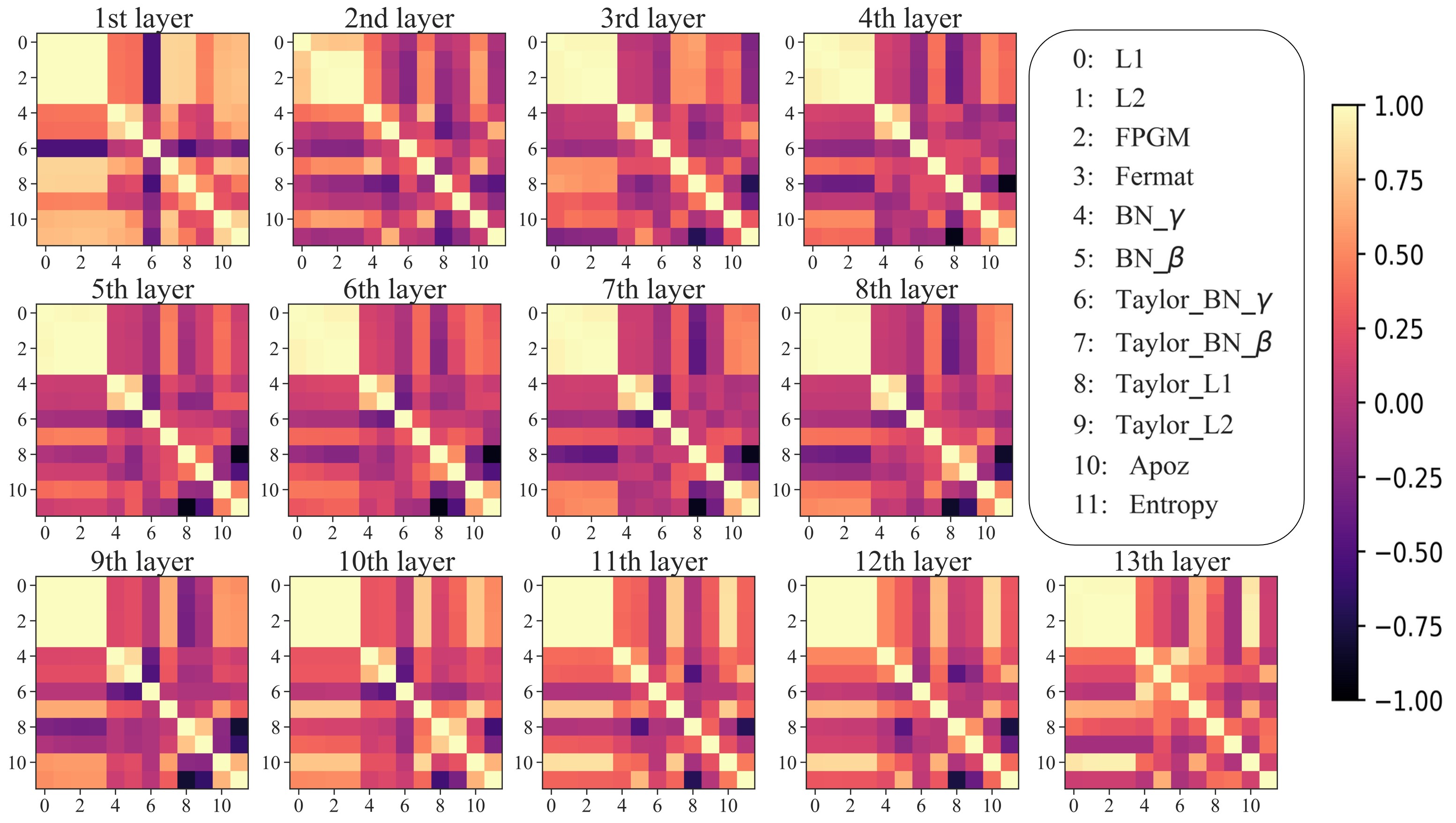}

\caption{The Spearman’s rank correlation coefficient~(Sp) of 12 pruning criteria in VGG16~\cite{simonyan2014very}. The color of each pair of pruning criteria represents the value of Sp, and the lighter the color~(close to 1), the stronger similarity.}
\label{fig:sp-1}
\end{figure*}

Conventional pruning methods mainly concentrate on designing a better criterion to increase the viable prune ratio without harming the performance, but few of them have introspected the actual correlations among them. Recent work~\cite{huang2020convolution} reveals that some of the filter pruning criteria, \textit{e.g.}, L1-Norm~\cite{li2016pruning}, L2-Norm~\cite{frankle2018the}, FPGM~\cite{he2019filter} and Fermat~\cite{huang2020convolution}, have a substantial similarity on the ``importance'' index of pruned filters in most layers. That is, some pruning criteria incline to remove alike filters, though their considerations are from a different perspective. From this motivation, we further extend the comparison to more criteria. In Figure~\ref{fig:sp-1}, we demonstrate the filters rank similarity under assorted state-of-the-art criteria and their variants using the Spearman's correlation coefficient~(Sp)~\cite{sedgwick2014spearman}, which is a non-parametric correlation measurement of rankings, and able to assess the relationship between two variables using a monotonic function. Specifically, for two sequences $X = \{x_1,x_2,\cdots,x_n\}$ and $Y = \{y_1,y_2,..,y_n\}$, if the rank of $X$ and $Y$ are $\{x_1\prime,x_2\prime,\cdots,x_n\prime\}$ and $\{y_1\prime,y_2\prime,\cdots,y_n\prime\}$ respectively, the definition of the Sp between them is
\begin{equation}
    \rho(X,Y) = 1 - \frac{6\sum_{i=1}^Nd_i^2}{n(n^2-1)},
    \label{equ:sp}
\end{equation}
where $d_i = x_i\prime - y_i\prime$. In general, if the Sp is above 0.8, there is a strong confidence that the two sequences of variables are highly similar. Figure~\ref{fig:sp-1} indicates an empirical fact that although each criterion is competent for filter pruning, the Sp correlation on their measured ``importance'' rank has a discrepancy. 
For example, in the first layer, Taylor BN\_$\gamma$~\cite{molchanov2019importance} has negative Sp value with other criteria, \textit{i.e.}, it tends to remove different filters comparing to other criteria. Additionally, in the intermediate layers, this discrepancy becomes obvious on more criteria, excluding the four criteria mentioned in \cite{huang2020convolution}. These inconsistencies imply that some of the pruning criteria may prune the potentially significant filters, \textit{e.g.}, a filter may be viewed as an important filter by one pruning criterion whereas another criterion may judge it unnecessary. Hence, each criterion may only be a partial view of the comprehensive ``importance'' of filters. Thus, inspired by this phenomenon, we consider a problem: can we introduce one criterion that integrates both the criteria diversity and their advantages as much as possible?

To solve this problem, we propose a novel framework to layerwise integrate the existing filter pruning criteria by examining the criteria diversity on their ``importance'' measurements, filters ``importance'' rank, and the discrepancy on their similarity. The proposed framework contains two stages: Criteria Clustering and Filters Importance Calibration. 
Since the searching space of the candidate combination pools will be extremely large, especially when the network is particularly deep, \textit{e.g.}, ResNet152 has over a hundred layers, finding the solution for the blending criterion among them is non-trivial. Therefore, to ease the above problem and to ensure the variance on blending candidates, we condense the pruning criteria via layerwise clustering based on the rank of ``importance'' score during the first stage. Then, in each cluster, we propose a calibration factor to adjust their significance for the ensemble. In the second stage, we introduce an Evolutionary Algorithm to optimize the combination of the calibration factor and the blending candidates sampled from each cluster and ultimately generate the optimal blending criterion. Our contributions are summarized as follows:

\begin{itemize}
\item To the best of our knowledge, our work is the first to explore the correlation among some existing filter pruning criteria and propose a simple-yet-effective ensemble framework to integrate different pruning criteria and generate an optimal criterion.
\item Compared to the conventional filter pruning methods, our method searches for the optimal pruning criteria automatically without manual interaction and empirical knowledge prior. The comprehensive experiments on benchmarks exhibit that our method can further outperform the current state-of-the-art methods. Especially, for ResNet56 pruning in CIFAR-100, our pruned model can exceed the original model performance.

\end{itemize}

\begin{figure}[tbp]
\begin{center}
\centering

\includegraphics[width=1.05\linewidth]{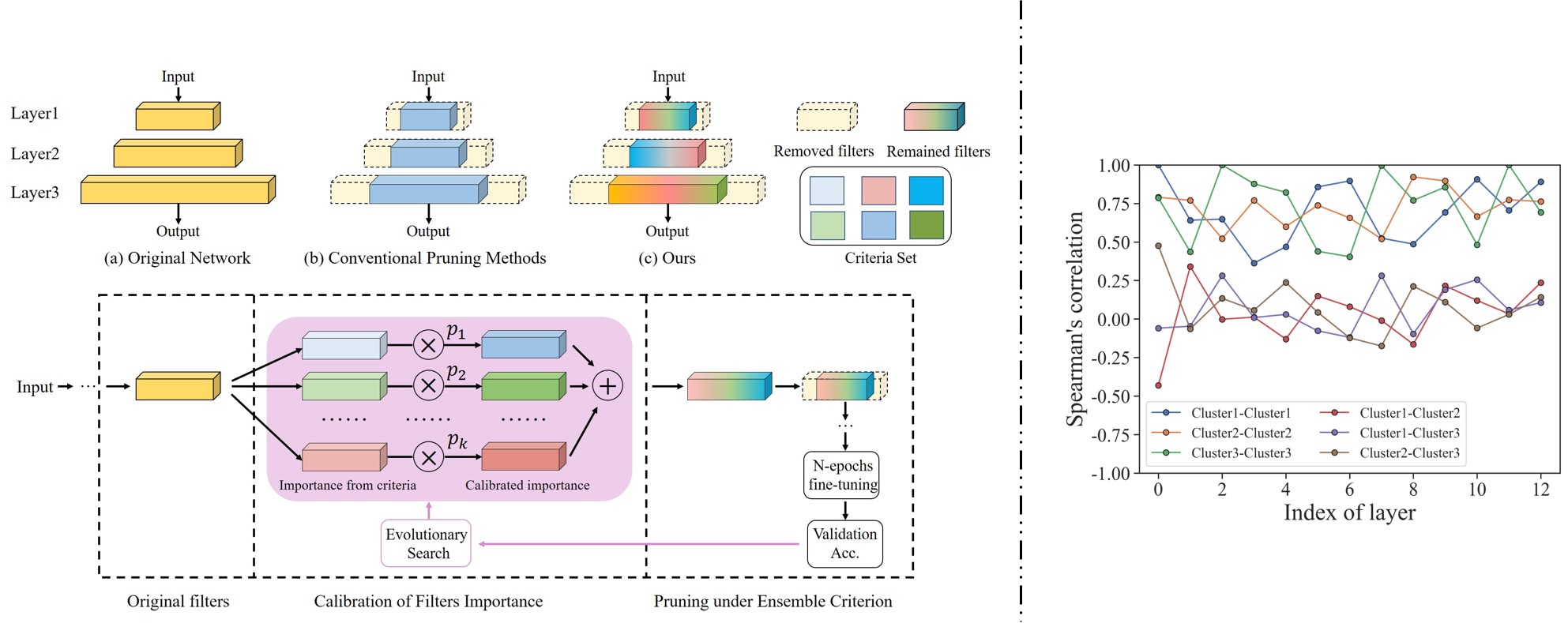}

\end{center}
\caption{\textbf{Left}: Overview of our framework for pruning. (a)-(c) show three layers within the original unpruned network, the pruned network under certain conventional filter pruning criterion~(denoted in blue) and the pruned network under the blending criterion. The three-step blending process in one layer is illustrated in the dash-line bounding box, where the notation $``\oplus"$ denotes the element-wise addition and the notation $``\otimes"$ denotes the multiplication between each filter score and the corresponding calibration factor $p$. \textbf{Right}: Average Sp of the criteria score between two clusters.}
\label{fig:structure}
\end{figure}

\section{Related Works}\label{related}

% \paragraph{Filter Pruning.} 
\subsubsection{Filter Pruning} 
To perform fast inference without largely sacrificing the accuracy, filter pruning focuses on removing the insignificant filters from the original network~\cite{luo2017thinet,guo2020multi}. Therefore, it is crucial to construct an effective criterion to evaluate the ``importance'' or contribution of the filters. Previously proposed criteria can be mainly categorized into four folds: (1) Filter-based criteria, which consider the property among filters as the importance indicator. The scoring metrics can be the filter's L1-Norm~\cite{li2016pruning} or L2-Norm~\cite{frankle2018the}. The underlying assumption of these methods is that the small norm parameters play a less informative role on the final prediction. Recent FPGM~\cite{he2019filter} and Fermat~\cite{huang2020convolution} advance the local filter property for the correlation of multiple filters in a layer. (2) Batch Normalization (BN) based criteria, which estimate channel importance via the value of the parameter inside each BN layer. Network Slimming~\cite{liu2017learning} and SSS~\cite{huang2018data} consider taking the magnitude of the scaling factor $\gamma$ to reveal the importance of their corresponding filter. (3) Feature maps activation based criteria, which take the activation value in each feature map as a proxy for its importance, for instance, the average percentage of zeros (APoZ) and the  in-between entropy on the channel maps~\cite{hu2016network,luo2017entropy}. (4) First-order Taylor based criteria, which estimate the filter's contribution with respect to the cost function and the scoring function is designed based on the Taylor expansion~\cite{molchanov2016pruning}.
One recent work is proposed to adaptively select pruning criterion for each layer to better suit the filter distribution across layers~\cite{he2020learning}. Our framework differs from~\cite{he2020learning} in two folds: (1) for each layer, we integrate different criteria based on the similarity of pruned filter selection, while~\cite{he2020learning} used only one pruning criterion for each layer; (2) according to the findings on~\cite{huang2020convolution} and the experiments (see Figure~\ref{fig:sp-1}), for L1-norm~\cite{li2016pruning}, L2-Norm~\cite{frankle2018the} and FPGM~\cite{he2019filter} used in~\cite{he2020learning}, these three criteria tend to remove the consistent set of filters in most layers. Thus, although one of them is selected, the strength of different criteria are not sufficiently utilized.

\subsubsection{Evolutionary Algorithms}
Recently, the Evolutionary Algorithms~(EA)~\cite{beyer2002evolution} and their variants are widely used in Network Compression and Neural Architecture Search~(NAS) areas as the EA can flexibly solve the multi-objective optimization problem and combinatorial optimization with conflicting objectives. In Network Compression, MetaPruning \cite{liu2019metapruning} applied an evolutionary search to find the high accuracy pruned network under the soft or hard constraints. \cite{junior2019pruning} leveraged an Evolution Strategy~(ES) algorithm to find a good solution for multi-objective optimization problems. In NAS, \cite{real2017large} modified the EA to search for high-performance neural network architectures for large and realistic classification dataset.~\cite{real2019regularized} introduced the novel aging evolution such that the tournament selection can be biased to choose the younger genotypes. In this paper, our target is totally different from all the previous work utilizing the EA. Our work leverages the EA to search for the blend of filter pruning criteria and our empirical results demonstrate the effectiveness.

\section{Proposed Method}\label{proposed}

% \begin{minipage}{0.8\linewidth}
\begin{algorithm}[!t]  
    \caption{Framework for Blending Pruning Criteria}\label{alg:all_in_one}   
    \begin{multicols}{2}
    \textbf{Input:} Unpruned model $\phi$ with $L$ convolution layers; Criteria Set $U_{criteria} = \{f_i\}_{i=1}^I, I$ is the number of included criteria; Convolution filters in layer $l$: $F^l = \{\mathbf{F}_i^l\}_{i=1}^{\lambda_{l}}, l=1,2,\cdots,L$; Number of clusters each layer: $K$;
    
    \textbf{Output:} Pruned model $\phi_\text{blended}$ over the optimal integrated criterion $S_\text{blended}$.
        
    \begin{algorithmic}[1]
    \State \algorithmiccomment{Calculation of Importance}
    \For{$l$ from 1 to $L$} 
        \For{$i$ from 1 to $I$}  
        \State$\mathbf{S}_i^{l} \gets f_i(\mathbf{F}_1^l,\mathbf{F}_2^l,\cdots,\mathbf{F}_{\lambda_{l}}^l)$ 
        \EndFor  
        \State $S^l_{Criteria} \gets$  \{${\mathbf{S}_1^{l},\mathbf{S}_2^{l},\cdots,\mathbf{S}_I^{l}}$\}
    \EndFor 
    
    \State \algorithmiccomment{Clustering Criteria}
        \For{$l$ from 1 to $L$}  
            \For{i from 1 to $I$} 
            
            \State\algorithmiccomment{ Calculate the Spearman’s correlation matrix}
            \State $Sp^l_{ij} \gets \rho(\mathbf{S}_i^{l},\mathbf{S}_j^{l}), j=1,\cdots,I$
            \EndFor  
            \State$(C^l_1,\cdots,C^l_K) \gets$  KM($Sp^l_{1:},\cdots,Sp^l_{I:}$)
        \EndFor  
    \State Obtain K-Means results: $(C^:_1,\cdots,C^:_K)$        
  
    \State \algorithmiccomment{Evolutionary Search}
    \State \textbf{EA Hyperparameters:} Population size $\mathcal{N}$, Number of iterations $\mathcal{I}$, Mutation Probability $\mathcal{M}$, Crossover Probability $\mathcal{C}$, Drop Ratio $\mathcal{D}$, Finetuned Epochs $\mathcal{E}$.    
  
    \State $\mathcal{G}_0 \gets (\mathbf{P}_0,\mathbf{S}_0)_{\mathbf{P}_0 \sim \mathcal{U}[0,1],\mathbf{S}_0 \sim (C^:_1,\cdots,C^:_K)}$
    \State $\mathbf{P}_0,\mathbf{S}_0 \in \mathbb{R}^{\mathcal{N}\times{L}\times{K}}$
    \For{$iter$ from 1 to $\mathcal{I}$}
    \State $\mathcal{G}_{Crossover} \gets $ Crossover($\mathcal{G}_{iter-1}$, $\mathcal{C}$)
    \State $\mathcal{G}_{Mutation} \gets $ Mutation($\mathcal{G}_{Crossover}$, $\mathcal{M}$)
    \State $\mathcal{G}_{Drop} \gets $ Drop($\mathcal{G}_{Mutation}$, $\mathcal{D}$)
    \State $\mathcal{G}_{iter} \gets \mathcal{G}_{Drop}$
    \State $\phi_{iter} \gets$ Blending($\mathcal{G}_{iter}$)
    \State $\text{Acc}_{\phi_{iter}} \gets$ Finetune($\phi_{iter}$,$\mathcal{E}$)
    \EndFor
    \State $\mathbf{P}_{topk},\mathbf{S}_{topk}\gets$ TopK($\text{Acc}_{\phi_{\mathcal{I}}}$)
    \State $\phi_\text{blended},S_\text{blended} \gets$ Final($\mathbf{P}_{topk},\mathbf{S}_{topk}$)
    \State \Return $\phi_\text{blended},S_\text{blended}$
    \end{algorithmic} 
    % }
    \end{multicols}
\end{algorithm}  
% \end{minipage}
In this section, we introduce our proposed framework for filter pruning. Given a CNN, our method adaptively generates an integrated criterion to identify the model redundancy layerwise. 
The proposed method consists of two stages: in the first stage, we divide different pruning criteria via clustering. In the second stage, we propose the calibration factors to combine criteria sampled from each cluster. Furthermore, the heuristic Evolutionary Algorithm~(EA) is applied to optimize the calibration factors and to search the optimal combination of criteria. The details of these two stages are illustrated in Algorithm~\ref{alg:all_in_one}. For notation, suppose that we are given $I$ criteria, \textit{e.g.}, L1-Norm, L2-Norm and FPGM, and the overall criteria set is denoted as $U_{criteria} = \{f_i\}_{i=1}^I$, where $f_i$ is the mapping to filter importance score under $criterion_i$. Consider a $L$-layer network, the filters set in $l$-th convolution layer is denoted as $F^l = \{\mathbf{F}^l_i\}_{i=1}^{\lambda_{l}}$, where $\lambda_{l}$ is the number of filters in layer $l$. The filters importance score $\mathbf{S}_i^{l}\in [0,1]^{\lambda_{l}}$ under $criterion_i$ is calculated by $f_i(\mathbf{F}_1^{l},\cdots,\mathbf{F}^{l}_{\lambda_{l}})$, where $i=1,\cdots,I$. Each component of $\mathbf{S}_i^{l}$ represents the importance score of the corresponding filter given by the $criterion_i$. For each criterion, the larger the score, the more important the filter is.  

\subsection{Criteria Clustering} 
We first consider the selection complexity for blending $K$ candidate criteria among $N$ given criteria on a $L$-layer CNN. When $N=12,K=6$, and $L=13$, we have ${(C_{12}^6)}^{13} \approx3^{38}$ combinations. For the commonly used model, the number of selection will be extremely large. In addition, from Figure~\ref{fig:sp-1}, we observe that some of the filter pruning criteria have a strong similarity on the rank of the criteria score. As a result, they tend to prune a similar set of filters in one convolution layer. 
Moreover, in traditional implementation of the ensemble method~\cite{dietterich2000ensemble,zhou2012ensemble}, its capability of achieving greater performance than an individual method comes from the diversity and effectiveness of the candidate methods that will be integrated. Given these above points, we cluster the given criteria set in each layer based on the Spearman's correlation matrix $\mathbf{Sp}$ before the blending, which is able to decrease the search space on the criteria selections efficiently. 
When the $K$ candidate criteria are obtained from each cluster in one layer, using the rule of product and Arithmetic Mean-Geometric Mean Inequality, the upper bound of the search space is $\left(\frac{N}{k}\right)^k$. And $%
    \left(\frac{N}{k}\right)^k = \prod_{j=0}^{k-1}\left(\frac{N}{k}\right) \leq \prod_{j=0}^{k-1}\left(\frac{N-j}{k-j}\right) = \frac{\prod_{j=0}^{k-1}(N-j)}{k!}=C_N^k.$

When $k \in [1,N]$ is selected appropriately, we have $\left(\frac{N}{k}\right)^k \ll C_N^k$. 
If we sample the candidate criteria from different clusters, the Sp value between them should be relatively small~(as shown in Figure~\ref{fig:structure}), \textit{i.e.}, their filter importance rank would be dissimilar. Therefore, this clustering of criteria can not only maintain the criteria diversity but also it can reduce the search space by selecting the number of clusters. Before clustering, we assign $criterion_i$ with a correlation vector, 
\begin{equation}
% \resizebox{.5\hsize}{!}{$%
Sp^l_{i:} = (\rho(\mathbf{S}_i^{l},\mathbf{S}_1^{l}), \rho(\mathbf{S}_i^{l},\mathbf{S}_2^{l}), \cdots, \rho(\mathbf{S}_i^{l},\mathbf{S}_I^{l})), 
% $%
% }
\end{equation}

where $ i=1, 2, \cdots, I$ and $\rho$ is Spearman's correlation defined in Eq.~\ref{equ:sp}. Subsequently, we conduct K-Means to cluster the correlation vectors of the criteria into $K$ clusters and obtain $K$ clustering sets $\{C^l_1,\cdots,C^l_K\}$. The two criteria in the same cluster have similar correlation vectors in the sense of Euclidean distance. We want to point out that the two criteria in the same cluster have large Sp correlation value~(see Figure~\ref{fig:structure}), which indicates the rank of these two criteria is similar. Therefore, we are able to sample criteria from each cluster whose in-between Sp is relatively small and indicate the adequate diversity of basic criteria for the ensemble.

\subsection{Filters Importance Calibration} 
After criteria clustering, we obtained $K$ clustering sets $(C_1^{l},C_2^{l},\cdots,C_K^{l})$ in layer ${l}$. To filter out the similar criteria, we sample the distinctive criterion score $\mathbf{S}_{i_k}^{l}$ from each cluster $C_k^l$ as the candidate for ensemble, where $i_k \in \{1,2,\cdots, I\}$, $k=1,2,\cdots,K$. Thus, to combine those selective criteria and integrate their importance measurements to identify the redundancy, we calibrate their filters importance with the introduced filter importance calibration factors $p_k^{l} \in [0,1]$. When filters in different layer extract multi-level features, we conduct layerwise criteria blending to adaptively discover their importance $S_{\text{blended}}^{l}  = \sum_{k = 1}^{K}p_k^{l} \mathbf{S}_{i_k}^{l}$.
The larger the value of $p_k^{l}$ reveals that the cluster $k$ is much significant for pruning filters in this layer. 
We denote $\phi(\{S_{\text{blended}}^{l}\}_{l=1}^L,N)$ as the network $\phi$ after finetuning $N$ epochs and discarding filters according to the ensemble score. As the pruning objective is to remove redundancy without harming the model performance too much, therefore, our framework for filter pruning tends to discover the blending criterion, such that its pruned network maximizes the accuracy after $N$-epoch finetuning in the validation set, %\textit{i.e.},
\begin{equation}
\resizebox{.8\hsize}{!}{$%
\max _{\left\{p_{k}^{l},\mathbf{S}_{i_k}^{l}; k=1, \cdots, K, l=1, \cdots, L\right\}} \text { Accuracy }\left(\phi\left(\left\{S_{\text {blended }}^{l}\right\}_{l=1}^{L}\right),N\right).
$%
}
\label{eqn:objective}
\end{equation}

Since the objective function~(\ref{eqn:objective}) is not differentiable, we consider using the Evolutionary Algorithm~(EA) oriented by the validation accuracy to optimize it, and the superiority of EA has been mentioned in the related works section. In the evolutionary search, the optimization fitness is the model evaluation result after $N$-epochs finetuning over part of the training set. The validation set and train set are split from the original training set, the splitting details are discussed in 
the experiments section. To be specific, each evolution gene consists of the calibration combination and the pruned network in terms of the corresponding calibrated criterion. All calibration factors are initiated from the uniform distribution $\mathcal{U}[0,1]$. Though the criteria in each cluster possess high similarity, they also have their ability to probe the network redundancy individually. To avoid sticking to one criterion that gives high ensemble accuracy, during crossover in EA, we give other criteria in the same cluster the opportunity to be selected again via random sampling in each evolutionary process.

After iterations of crossover, mutation, and drop, the TopK genes with the highest validation accuracy are considered the potential optimal calibrated pruning criterion. Then, under these blending results, we can obtain the gene with the highest accuracy after the final one-shot pruning and finetuning, and the criterion under this gene is considered the optimal filter importance measurement to discard unimportant filters in this architecture. 

\section{Experiments}\label{exp}

\begin{table}[!t]
  \centering
  
  \caption{Quantitative results on CIFAR-100 dataset}
  
  \resizebox{\textwidth}{!}{%
  \begin{threeparttable}
    \begin{tabular}{c|l|cc||c|l|cc}
    \toprule
    Model & Criterion & Pruned/Finetuned Acc.(\%) & Acc.$\downarrow$(\%)  & Model & Criterion & Pruned/Finetuned Acc.(\%) & Acc.$\downarrow$(\%) \\
    \midrule
    \multirow{13}{*}{\rotatebox{90}{VGG16\tnote{*}}} & L1-Norm    & 15.76/71.29 & 0.93  & \multirow{13}{*}{\rotatebox{90}{ResNet56\tnote{$\dagger$}}} & L1-Norm    & 52.16/69.43 & 0.07  \\
          & L2-Norm    & 16.24/71.32 & 0.90  &       & L2-Norm    & 50.75/69.45 & 0.05  \\
          & Apoz  & 5.69/70.91 & 1.31  &       & Apoz  & 2.03/63.68 & 5.82  \\
          &  BN\_$\gamma$ & 15.87/71.26 & 0.96  &       & BN\_$\gamma$ & 29.35/69.39 & 0.11  \\
          & BN\_$\beta$ & 6.92/71.45 & 0.77  &       & BN\_$\beta$ & 22.44/69.32 & 0.18  \\
          & Entropy & 11.80/71.09 & 1.13  &       & Entropy & 17.37/69.14 & 0.36  \\
          & FPGM  & 15.91/71.37 & 0.85  &       & FPGM  & 51.20/69.50 & 0.00  \\
          & Fermat & 15.39/71.39 & 0.83  &       & Fermat & \textbf{51.45}/69.47 & 0.03  \\
          & Taylor L1-Norm & 1.29/70.35 & 1.87  &       & Taylor L1-Norm  & 15.77/69.27 & 0.23  \\
          & Taylor L2-Norm & 16.19/71.19 & 1.03  &       & Taylor L2-Norm & 39.22/69.23 & 0.27  \\
          & Taylor\_BN\_$\gamma$ & 7.46/71.19 & 1.03  &       & Taylor\_BN\_$\gamma$ & 30.41/69.34 & 0.16  \\
          & Taylor\_BN\_$\beta$ & 7.10/71.16 & 1.06  &       & Taylor\_BN\_$\beta$ & 25.42/69.34 & 0.16  \\
          & \textbf{Ours} & \textbf{16.67/71.68} & \textbf{0.54} &       & \textbf{Ours} & 40.38/\textbf{69.82} &  \textbf{-0.32} \\
    \bottomrule
    \end{tabular}%
            \begin{tablenotes}\footnotesize
            \item[*] VGG16 original Acc.: 72.22\%
            \item[$\dagger$] ResNet56 original Acc.: 69.50\%
        \end{tablenotes}
    \end{threeparttable}
    }
  \label{tab:CIFAR100}%
\end{table}%

We evaluate the effectiveness of our proposed method over different image classification benchmarks. 

\begin{figure}[t]
\centering
\includegraphics[width=\columnwidth]{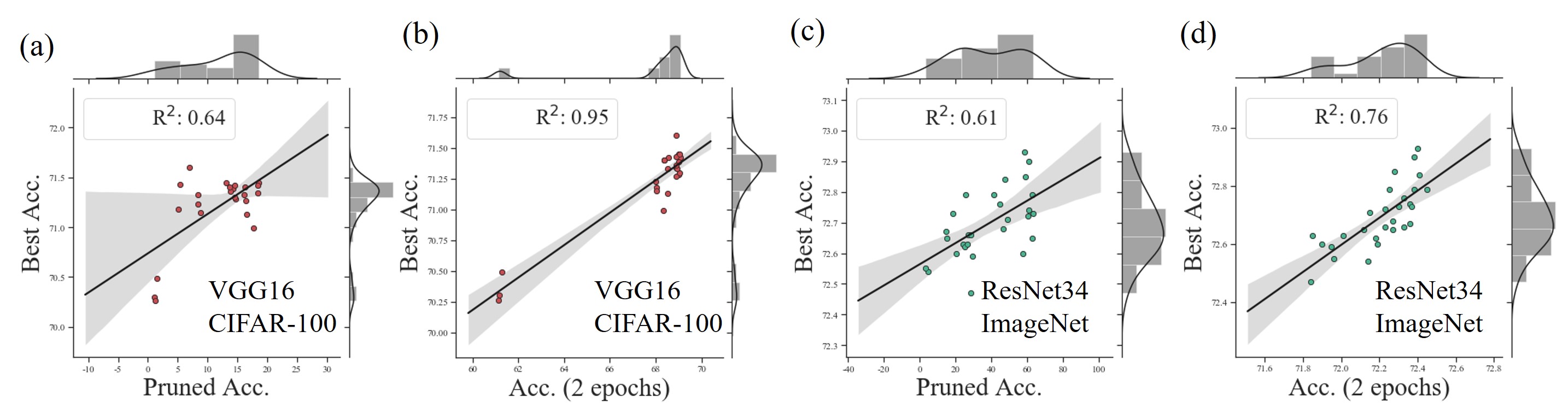}
\caption{The $R^2$ between Acc. on different finetuned epochs. (a) and (c): best finetuned Acc. vs pruned Acc.; (b) and (d): best finetuned Acc. vs Acc. after 2 epochs finetuning.}
\label{fig:tune}
\end{figure}

\begin{figure}[]
\begin{center}
\centering
\includegraphics[width=1\columnwidth]{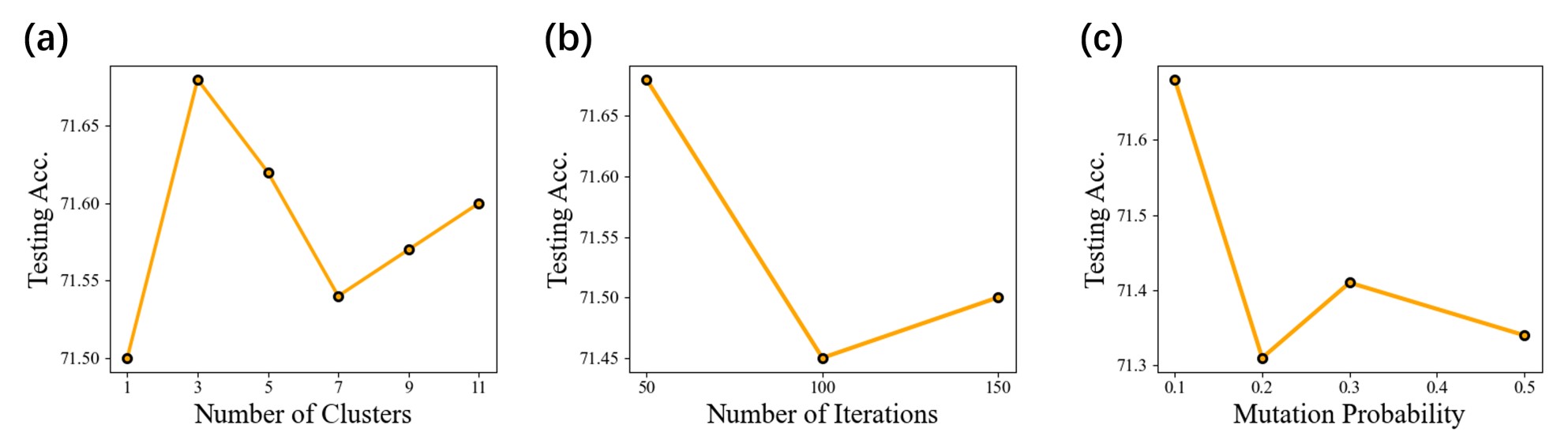}
\end{center}
\caption{The testing accuracy of VGG16 after finetuning. (a) Comparison of the performance under different number of clusters. Note that when the number of clusters equals to 1, we choose one criterion at each layer and when the number of clusters equals to number of criteria, we choose all criteria; (b) Comparison of the performance over different number of iteration in EA; (c) Comparison of the performance over mutation probability in EA.}
\label{fig:ablation}
\end{figure}

\subsubsection{Dataset} 
We conduct experiments on CIFAR-100~\cite{cifar} and ImageNet~\cite{ILSVRC15} datasets. CIFAR-100 has 50k train images and 10k test images of size 32 by 32 from 100 classes. 10\% of train images are split for validation and the remaining for training. ImageNet comprises 1.28 million train images and 50k validation images from 1000 classes. 50k out of 1.28 million train images (50 images in each class) are used for sub-validation. The cropping size 224 by 224 is used in our ImageNet experiments. Adopting the same predefined pruning configuration~\cite{li2016pruning}, we evaluate our method on VGG~\cite{simonyan2014very} and ResNet~\cite{he2016deep}.

\subsubsection{Results \& Analysis}
In Table~\ref{tab:CIFAR100} and Table~\ref{tab:imagenet}, we present the quantitative comparison on CIFAR-100 and ImageNet, where the average accuracy over three repeated experiments are attached (denoted as Acc.). Comparing to the baselines, our method performs the best in the same settings.

According to Figure~\ref{fig:structure}, the optimization fitness in the evolutionary search needs $N$-epochs finetuning over part of the training set. Is $N$-epoch finetuning necessary in evolutionary search? 

To illustrate, we take VGG16 on CIFAR-100 and ResNet34 on ImageNet as examples. First, we calculate the Pearson correlation coefficient~($R^2$) between the accuracy of the pruned model without finetuning and the best accuracy after completed finetuning. In Figure~\ref{fig:tune} (a) and (c), the value of $R^2$ indicates that the accuracy between the pruned model accuracy and the best accuracy does not have a strong linear relationship. Therefore, the accuracy of the pruned model without finetuning is not suitable as a metric for our Evolutionary process in Filters Importance Calibration. However, after several epochs of finetuning~(as shown in Figure~\ref{fig:tune} (b) and (d)), the $R^2$ improve significantly and it means that a certain amount of finetuning is necessary.

\begin{figure}
\begin{center}
\centering
\includegraphics[width=0.75\columnwidth]{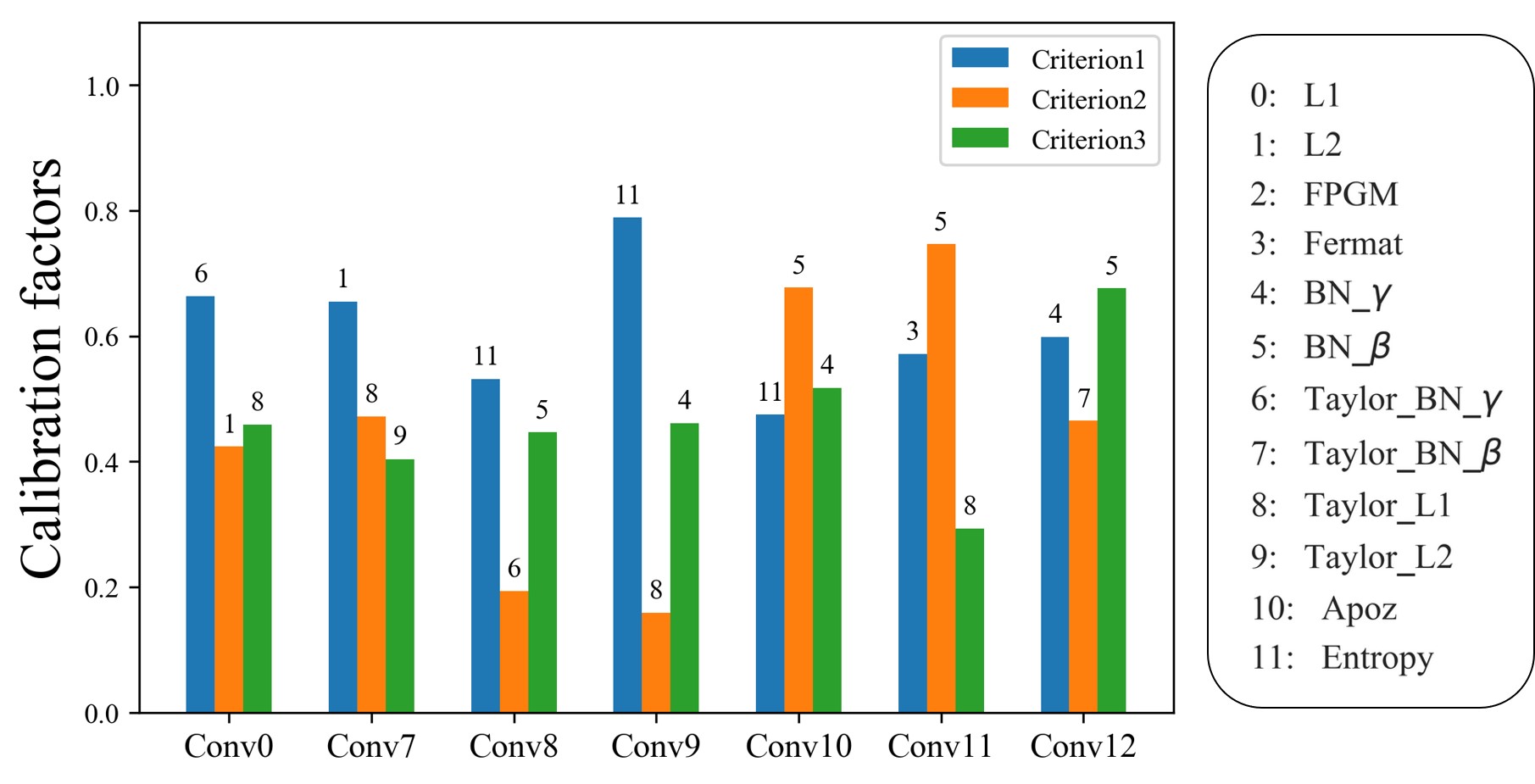}

\end{center}
\caption{The optimal value of calibration factors and the sampled criterion of each cluster at different layers of VGG16 on CIFAR-100. The annotation above each bar denotes the sampled criterion index.}
\label{fig:calibration2}
\end{figure}

From the clustering results, pruning criteria that have strong similarity will gather in the same cluster as expected. In Figure~\ref{fig:calibration2}, the value of calibration factors on each cluster are illustrated. Taking the Conv7 of VGG16 on CIFAR-100 as an example, Criterion1 that consists of L1-Norm, L2-Norm, and FPGM, is calibrated with the largest factor, which indicates that the first cluster contributes more significantly in this layer.

\section{Implementation Details}
For the normal training and the finetuning, we use the SGD optimizer with momentum and weight decay parameter 0.9 and 0.0001 respectively. For finetuning, the learning rate is started at 0.001. On CIFAR-100, we finetune the pruned model for 40 epochs in batch size 64. On ImageNet, we finetune for 20 epochs with a mini-batch size of 256.

For the evolutionary search setting, we consider the criteria as follows: L1-Norm, L2-Norm, FPGM, Fermat, BN\_$\gamma$ scale, BN\_$\beta$ scale, Entropy, Taylor L1-Norm, Taylor L2-Norm, Taylor BN\_$\gamma$, Taylor BN\_$\beta$. In each evolution iteration, we set the mutation probability to 0.1, crossover constant to 0.8, and drop probability to 0.05. In CIFAR-100, the population size is 20, the number of iterations is 50 and the drop ratio is 0.08. In ImageNet, the population size is 10, the number of iterations is 30 and the drop ratio is 0.1. After pruning the network based on the weighted sum of the scores, we finetune the pruned network on the validation split for 3 epochs in CIFAR-100 experiments and 1 epoch for ImageNet. Finally, the network is pruned under the optimal blending criterion obtained by EA, where the finetuning epochs are 40 on CIFAR-100 and 20 on ImageNet over the whole training set.

\begin{figure}[tbp]

\begin{minipage}[b]{0.48\linewidth}
    \centering
 \captionof{table}{ImageNet Quantitative Results}\label{tab:imagenet}
  \resizebox{\columnwidth}{!}{%
   \begin{threeparttable}
     \begin{tabular}{c|l|ccc}
    \toprule
    Model & Criterion & Pruned/Finetuned Acc.(\%)  & Acc.$\downarrow$(\%)  \\
    \midrule
    \multirow{13}{*}{\rotatebox{90}{ResNet34\tnote{*}}} 
          & L1-Norm    & 59.08/72.76  & 1.03  \\
          & L2-Norm    & 61.02/72.77  & 1.02  \\
          & Apoz  & 4.70/72.19  & 1.60  \\
          & BN\_$\gamma$ & 19.02/72.71  & 1.08  \\
          & BN\_$\beta$ & 4.12/72.59  & 1.20  \\
          & Entropy & 25.84/72.57  & 1.22  \\
          & FPGM  & \textbf{62.28}/72.78  & 1.01  \\
          & Fermat & 44.63/72.80  & 0.99  \\
          & Taylor L1-Norm & 25.71/72.67  & 1.12  \\
          & Taylor L2-Norm & 46.43/72.67  & 1.12  \\
          & Taylor BN\_$\gamma$ & 27.47/72.65  & 1.14  \\
          & Taylor BN\_$\beta$ & 18.00/72.67  & 1.12  \\
          
          & \textbf{Ours} & 61.33/\textbf{72.85}& \textbf{0.94}\\
          
          \bottomrule
    \end{tabular}%
    
         \begin{tablenotes}\footnotesize
        \item[*] ResNet34 original Acc.: 73.79\%
    \end{tablenotes}
    \end{threeparttable}
    }
\end{minipage}
\begin{minipage}[b]{0.48\linewidth}
    \centering
    \captionof{table}{Ablation on  
    Hyperparameters}\label{tab:ablation}
  \resizebox{\columnwidth}{!}{%
  \begin{threeparttable}
    \begin{tabular}{cccc}
    \toprule
    \#Clusters & \multicolumn{1}{l}{\#Iterations} & \multicolumn{1}{l}{Mutation Prob.} & \multicolumn{1}{l}{Fintuned Acc.} \\
    \midrule
    1     & 50    & 0.10  & 71.50 \\
    3     & 50    & 0.10  & 71.68\\
    5     & 50    & 0.10  & 71.62 \\
    7     & 50    & 0.10  & 71.54 \\
    9     & 50    & 0.10  & 71.57 \\
    11    & 50    & 0.10  & 71.60 \\
     3     & 100   & 0.10  & 71.45 \\
     3     & 150 & 0.10  & 71.50\\
     3     & 50 & 0.20  &71.31\\
     3     & 50 & 0.30  &71.41\\
     3     & 50 & 0.50  & 71.34 \\
    \bottomrule
    \end{tabular}%
    
    \end{threeparttable}
    }

\end{minipage}

\end{figure}

\section{Ablation Studies}

In this section, to understand the performance of our method in different settings, we conduct the following ablation experiments on the number of clusters used in Criteria Clustering and the hyperparameters of Filters Importance Calibration. The ablation results are shown in Figure~\ref{fig:ablation}. The search space of EA is related to the number of clusters. In each layer $l$, the larger the number of clusters $K^l$, the harder the optimization of the calibration factors $p^l$. From our experiments, the best performance appears when $K^l=3$ for VGG16. Also, we compare the performance with different hyperparameters of EA, including the number of evolution iterations and mutation ratio. As we see in Figure~\ref{fig:ablation}(b), the increment of iterations is not sufficient for the increment of performance. Therefore, we choose the relatively small number of iterations and use the top-K strategy.  From Figure~\ref{fig:ablation}(c), we observe that the large mutation probability may harm the search. The ablation study on the EA hyperparameters and the number of the cluster are shown in Table~\ref{tab:ablation}.

\section{Conclusion}

In this paper, we propose a novel framework for filter pruning. In the first stage, we reduce the searching space for the criteria selection and fit the requirement for the ensemble using clustering. In the second stage, the criteria blending problem is formulated as an optimization problem on filters' importance calibration. The comprehensive experiments on various benchmarks exhibit that our blended criterion is able to outperform the current state-of-the-art criteria. Besides, we explore the correlation among existing filter pruning criteria and provides a way to obtain effective criteria without manual efforts. 

\section{Acknowledgments}
S. L. and H. Y. were partially supported by the NSF grant DMS-1945029 and the NVIDIA GPU grant.

% \bibliography{./ref}

\end{document}